\documentclass[journal]{IEEEtran}
\usepackage{graphicx}
\usepackage{amsmath}
\usepackage{amsthm}
\usepackage{bm}
\usepackage{amsmath,amssymb,amsfonts}
\usepackage{algorithmic}
\usepackage{graphicx}
\usepackage{textcomp}
\usepackage{xcolor}

\usepackage{times}
\usepackage{helvet}
\usepackage{courier}
\begin{document}
%
\title{College Student Retention Risk Analysis From Educational Database using Multi-Task Multi-Modal Neural Fusion}
\author{
Mohammad Arif Ul Alam\\
Department of Computer Science, University of Massachusetts Lowell\\
Lowell, MA, USA\\
mohammadariful\_alam@uml.edu
}
\maketitle
\begin{abstract}

We develop a Multimodal Spatiotemporal Neural Fusion network for Multi-Task Learning (\emph{MSNF-MTCL}) to predict 5 important students' retention risks: future dropout, next semester dropout, type of dropout, duration of dropout and cause of dropout. First, we develop a general purpose multi-modal neural fusion network model \emph{MSNF} for learning students' academic information representation by fusing spatial and temporal unstructured advising notes with spatiotemporal structured data. \emph{MSNF} combines a Bidirectional Encoder Representations from Transformers (BERT)-based document embedding framework to represent each advising note, Long-Short Term Memory (LSTM) network to model temporal advising note embeddings, LSTM network to model students' temporal performance variables and students' static demographics altogether. The final fused representation from \emph{MSNF} has been utilized on a Multi-Task Cascade Learning (\emph{MTCL}) model towards building \emph{MSNF-MTCL} for predicting 5 student retention risks. We evaluate \emph{MSNF-MTCL} on a large educational database consists of 36,445 college students over 18 years period of time that provides promising performances comparing with the nearest state-of-art models. Additionally, we test the fairness of such model given the existence of biases.
\end{abstract}
\section{Introduction}
The U.S. National Center for Education Statistics (NCES) reports that in United States, the average retention rate for higher education institutions is 71\% \cite{nces}. While, 57\% of college admitted students do not complete four-year colleges within six years, 33\% of them drop out from college without any degree \cite{nces}. For some students, dropping out is the culmination of years of academic hurdles, missteps, and wrong turns. For others, the decision to drop out is a response to conflicting life pressures, the need to help support their family financially or the demands of caring for siblings or their own child. Dropping out is sometimes about students being bored and seeing no connection between academic life and "real" life. It's about young people feeling disconnected from their peers and from teachers and other adults at school \cite{reason}. Although the reasons for dropping out vary, the consequences of the decision are remarkably similar. Low retention rates not only impact the financial well-being of individuals but the economy as a whole, college dropouts are more likely to head down a path that leads to lower-paying jobs, poorer health, and the possible continuation of a cycle of poverty that creates immense challenges for families, neighborhoods, and communities \cite{nces}. Low retention rates also adversely affect the reputation of the educational institution and could lead to potential loss of funding and inability to compete for quality students \cite{robert}. Thus, improving student retention is of paramount importance at institutions of higher education.

Many researchers have proposed to model factors impacting student dropout from large scale educational database using statistical and machine learning models. Most researchers have focused on using static or temporal structured data, such as GPA, SAT scores etc., that are readily available in institutional databases \cite{PrenkajSM20}. Some of the researchers proposed to use unstructured text analysis such as advising notes, forum post, social media status, online chats and email mining using natural language processing techniques to predict student dropout \cite{drop6,drop7}. However, none of the researchers proposed to combine structured and unstructured data altogether in spatiotemporal fashion that can provide significant promise in this domain of research. We propose \emph{MSNF-MTCL} with the following {\bf key contributions}:
\begin{itemize}
  \item We develop a novel multimodal spatiotemporal neural fusion model \emph{MSNF} for educational database to fuse temporal student advising notes extracted BERT embedding, temporal student performance variables and static student demographic information via temporal document encoder, temporal performance encoder and static demographic encoder respectively.
  \item We develop a cascaded information network-based Multi-Task Cascade Learning (MTCL) layer on the top of the fusion layer to build our core \emph{MSNF-MTCL} model by placing lower-level tasks at earlier layers so that the features learned for these tasks may be used by higher-level tasks for 5-tasks MTCL problem.
  \item We evaluate \emph{MSNF-MTCL} on a large scale collected data from a University from a third world country via comparing the performance with nearest state-of-art solutions.
  \item Additionally, we tested the existence of biases and applied bias mitigation technique to confirm fairness of \emph{MSNF-MTCL}.
\end{itemize}

\section{Related Works}

Traditionally, education researchers run surveys to find the facts impacting dropped out students dropout that include academic difficulty, adjustment problems, lack of clear academic goals, lack of commitment, inability to integrate with the college community, uncertainty, incongruence, isolation as factors involved in student dropout \cite{drop7}. The surveys result some key factors such as past and current academic success, high school GPA, SAT scores \cite{drop8}, major and number of credit hours taken during the first semester \cite{drop9}. effect of financial aid \cite{drop10}. Machine learning techniques on educational database has been relatively new \cite{drop1,drop2,drop3,drop4,drop5}. Perez et. al. proposed logistic regression and decision tree based dropout prediction from static students' data \cite{drop1}. \cite{drop2} proposed a link-based cluster ensemble for predicting student dropout from mixed-type (categorical and continuous) educational dataset. \cite{drop3} presented benchmark student dropout definition and dropout prediction paradigm by developing machine and deep learning techniques and their related privacy concerns from static and temporal structured data. \cite{drop4} proposed logit leaf model (LLM) on students classroom characteristics, cognitive and behavioral engagement variables and other static variables available from online students' enrollment database. \cite{drop5} proposed a Generalized mixed-effects random forest model to analyze hierarchical data to predict engineering students' dropout from static data from large scale educational dataset. On the other hand, student dropout prediction from advising notes has been explored only once \cite{drop6} that proposed a sentiment analysis technique to mine advising notes towards predicting students' dropout. Additionally, this paper proposed an explanation i.e. weighted ranking of contributing sentiments towards predicting students' dropout. \cite{yu21} proposed a fair student dropout prediction system from educational database. \cite{PrenkajSM20} analyzed the challenges of student dropout from static database that involves definition, machine learning techniques to be used, evaluation measures and privacy concerns.

Combining structured and unstructured data has been popular in image processing content learning and electronic health record analytics for decades. \cite{wan14} proposed deep spatial CNN model to extract features from image-text pairs. \cite{TamGSVFCCMWMB21} presents LSTM-CNN fusion to combine clinical image and electronic health records together for predicting clinical events derived cohorts. \cite{Wu21} presents utilized unstructured-structured text fusion model for predicting cognitive engagement. Similar approach has been conducted in many domains such as mortality prediction \cite{BaxterKSYHN20}, structured visualization from unstructured texts \cite{LiFLLZ21}, financial transaction prediction \cite{AuAK21} and so on.

Multi-task learning (MTL) has been investigated mostly by computer vision researchers that are categorized in many terms such as shared trunk, cross-talk, prediction distilation, task routing. In NLP, the MTL falls under many categories. Traditional feed-forward neural networks (non-attention based) focused on developing structural resemblance of shared global feature extractor followed by task-specific output branches where features are word representations \cite{CollobertWBKKK11}. Recurrent neural network models in MTL mostly focused on novel recurrent neural architectures adopted in multi-task fashion with multi-variant parameter sharing schemes i.e., one-to-one, one-to-many and many-to-many or task specific LSTMs \cite{LiuLWT15,DongWHYW15}. Cascaded information techniques mostly focused on lower-level tasks at earlier layers so that the features learned for these tasks may be used by higher-level tasks \cite{cascaded2019}. Adversarial feature separation techniques introduce an adversarial learning framework for MTL in order to distill learned features into task-specific and task-agnostic subspaces. Their architecture is comprised of a single shared LSTM layer and one task-specific LSTM layer per task\cite{RuderBAS19}. BERT in MTL mostly focused on adding shared BERT embedding layers on the traditional, LSTM or cascaded information technique \cite{LiuWJCZAHCPCG20}.

To our best knowledge, \emph{(MSNF-MTCL)} is the first of its kind, that develops a Multimodal Spatiotemporal Neural Fusion for MTL model combining structured, unstructured, spatio-temporal contexts altogether on educational data. More elaborately, we design multimodal neural network model to fuse static students' structured demographic information, temporal students' structured performance information and temporal students' unstructured advising notes altogether and develop a novel classification model towards predicting student dropout, next semester dropout and dropout cause identification.

\section{Data Description}
We obtained an educational database from a private university located in a third world country consists of 36,445 undergraduate students where female (10,237) and male (26,208) students' ratio ( 28\% by 72\%) is similar to national literacy statistics of the country. Among the students, 14\% are dropped out (female-male dropout are 11\% and 15\%) in any point of their study. While any dropout incident happened, dropped out students were contacted by university counselling office via phone to analyze the incident which has been categorized into two classes (1) temporary dropout, (2) permanent dropout. Here, the causes of permanent dropout has been sub-categorized into 10 classes (financial, family, marriage, sickness and so on) and temporary dropout has been sub-categorized into 14 classes (financial, internship, sickness, accident, marriage, COVID-19 related, family member death, struggling with grades and so on). Both of temporary and permanent dropout causes have 9 overlaps and in total 15 unique causes have been structured to represent any kind of dropout causes. It should be noted that location transfer and university transfer reasons were not considered as dropout in the inclusion criteria, and these information has been removed from every statistics. While getting admitted, students were provided few demographic data related to students personal profile, prior education details, family information and financial information. Since the admission, university administration has been recording students' temporal performances in each courses taken along with few administrative structured information such as payment due, blocked to register for next semester (due to any critical incidents, past significant payment dues), scholarship awarded etc. Each semester, students were required to visit to his/her academic advisor to discuss various topics related to academia which is more likely to be the first month of the semester. Sometimes, students were blocked from registering to next semester without consulting academic advisor due to many reasons, such as, poor grades, excessive missing of attendance, payment dues. However, students also could schedule meeting with their academic advisor anytime of the semester to discuss various topics (from personal to academic). It should be noted that, only primary cause of dropout has been noted during the counselling session. Table \ref{tab:description} and Table \ref{tab:features} present the details of the statistics of the dataset and features information derived/extracted from the database respectively.

\begin{table}[!h]
 \begin{center}
\begin{small}
 \caption{Description of the obtained educational database}
 \label{tab:description}

 \begin{tabular}{|p{0.8cm}|p{1.2cm}|p{1.3cm}|p{1.5cm}|p{1.5cm}|}
 
  \hline
 {\bf Gender}   & {\bf Count}  & {\bf Dropout} & {\bf Temporary} & {\bf Permanent} \\
 \hline
   
   Female & 5,498 (24\%)  & 1,103 (11\%) & 717 (65\%) & 386 (35\%) \\ \hline
   Male & 17,897 (76\%)  & 3,857 (15\%)& 2,931 (76\%)& 926 (15\%) \\ \hline
   
   Total & 23,395  & 4,960 (14\%) & 3,648 (74\%) & 1,312 (26\%) \\ \hline

 \end{tabular}
\end{small}
 \end{center}
\end{table}

\begin{table}[!h]
 \begin{center}
\begin{small}
 \caption{Description of the features provided/generated from the educational database}
 \label{tab:features}

 \begin{tabular}{|p{1.5cm}|p{6cm}|}
 
  \hline
 {\bf Variables }   & {\bf Features} \\
 \hline
   
   Static and structured Demographic & birth date, age, gender, religion, starting major, transferred credits, blood group, birth place, permanent address, local address, Secondary School grade, higher school grade, marital status, source of finance, part-full time, local guardian, parents financial income\\ \hline
   Temporal and structured Performance & new credits taken, credits retaken, passing credits, failed credits, overall attendance, average semester starting GPA, average semester GPA, average semester ending GPA, number of exams unattended since admission, number of exams unattended in this semester, number of counselling scheduled, amount of payment due in this semester, number of payment dues since admission, study duration, blocked from registering in next semester, number of block since admission, scholarship amount, accommodation status (on/off campus), total scholarship till date, average scholarship per semester\\ \hline
   Temporal Advising Notes & {\it structured:} reason of counselling visits, counselling conduct date, counselling result (no result or cause of dropout) {\it unstructured:} counselling note\\ \hline
   
   Dropout Causes & *financial, *family, *marriage, *physically ill, *death of family member, *personal, $\dagger$death, *accident, *struggling with grades, *COVID-19 family death, COVID-19 financial, COVID-19 online class attending hardship, internship, traveling, mentally ill\\ \hline

 \end{tabular}
\end{small}
 \end{center}
\end{table}

\begin{figure*}
 \centering
 \includegraphics[width=0.75\linewidth]{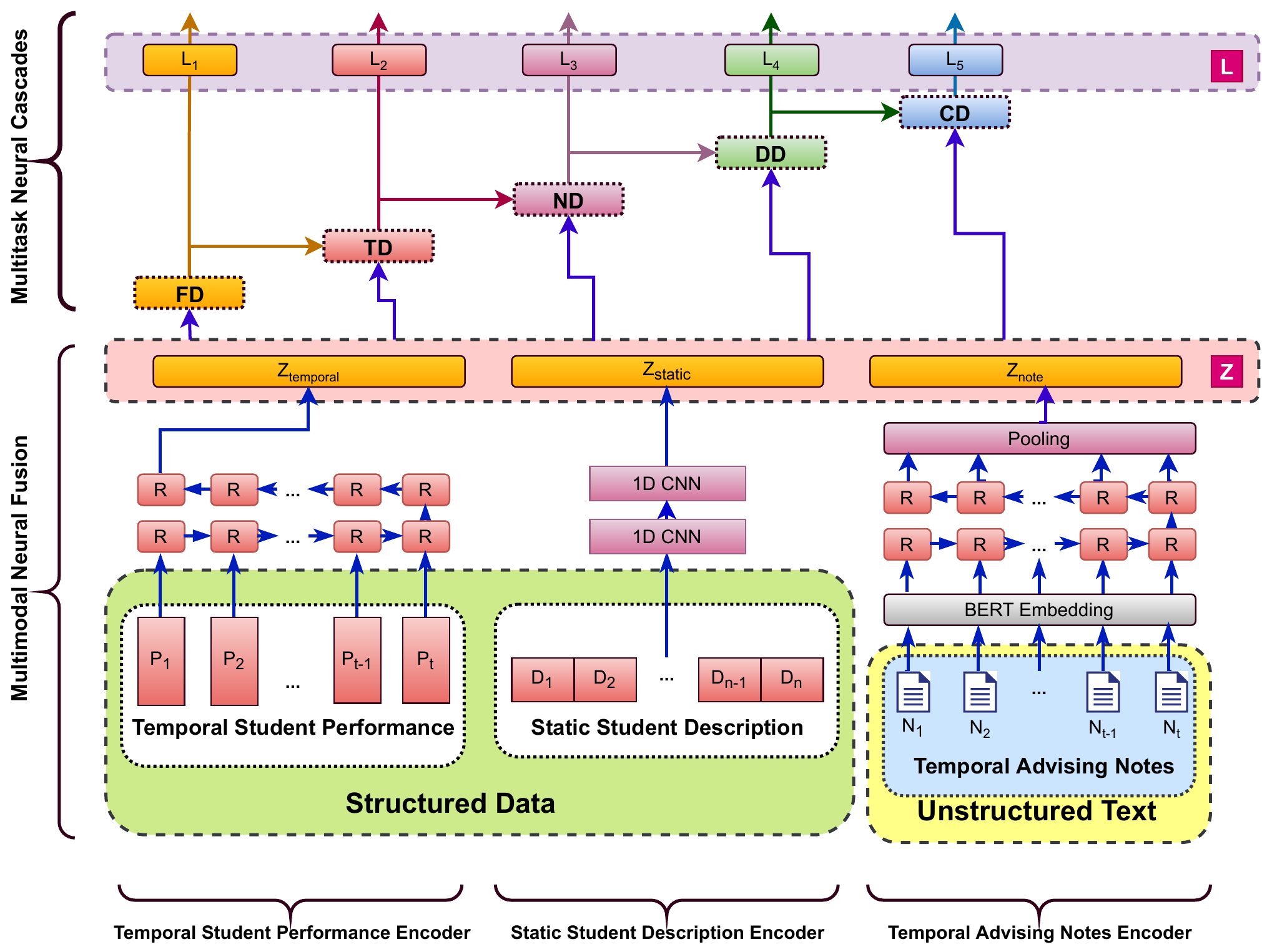}
 \caption{Overall Architecture of Multimodal Spatiotemporal Neural Fusion (MSNF) Network model for predicting student dropout risks i.e. dropout, next semester dropout and cause of dropout}
 \label{fig:EAAI_Overview}
\end{figure*}

\section{Multi-Task Multi-Modal Neural Fusion Model for Predicting Student Retention Risks}
In this section, we describe the problem formulation, multi-modal spatiotemporal neural fusion and multi-task neural cascade networks to solve student retention risks prediction. The overall framework has been shown in Fig~\ref{fig:EAAI_Overview}. The lower module "Multi-Modal Fusion" generates, $L$, a spatiotemporal fused layer that has been shared across all tasks, while the upper module "Multitask Neural Cascades" represent task-specific outputs, $L$, in our case $L\in \{L_1, L_2, L_3, L_4, L_5 \}$. 

\subsection{Multimodal Spatiotemporal Neural Fusion}
This module consists of advising note representation via BERT-based document embedding, sequential encoder network on temporal advising note documents from BERT embedding, development of temporal structured performance information encoder, development of static information encoder and a fusion layer that has been shared by each task of Multi-Task Cascade network. The input can be represented as $X \in \{ P, D, N\}$ where $P, D, N$ represent temporal structured performance data, static students demographic data and temporal students' advising/counselling notes. The output of this layer is fused representation of spatiotemporal inputs of $X$ which can be represented as $Z \in \{ Z_{temporal}, Z_{static}, Z_{note}\}$.

\subsubsection{Static Information Encoder ($Z_{static}$)}
The static student description (Table~\ref{tab:features}) data $D\in\{ D_1, D_2 .. D_n\}$ has been converted into one-hot vectors through static student description encoder to generate output $Z_{static}$. This encoder consists of a series of convolution (CNN) layers, where each CNN layer further followed by batch normalization, max pooling, and dropout layer. The first 1D CNN layer takes the one-hot encoded static feature and structured demographic data (size: 120) as input and performs the filter operation with 8 filters of size 11. The outputs of the first CNN layer are passed to the second CNN layer (16 filters with a size of 5). Next, the outputs of the second CNN layer are passed to the third CNN layer (32 filters with a size of 3). Finally, the summary of all the spatial features of a static input feature is passed to the flatten layer to produce a 1D feature vector of size 50.

\subsubsection{Temporal Student Performance Encoder ($Z_{temporal}$)}
To capture the longer dynamics in the temporal dimension of the temporal student performance data, $P \in \{ P_1, P_2 .. P_t\}$ where $t$ represents the time, we have used two consecutive LSTM layers: The first LSTM layer consists of 75 neurons, and the second one with 55 neurons. Each LSTM layers are followed by a dropout and batch normalization layer. Next, a dense layer of 50 neurons, followed by a dropout and batch normalization layer, is connected to another dense layer with 40 neurons. Finally, informative features of the input $P$ have been extracted to generate final encoded layer $Z_{temporal}$.

\subsubsection{Sequential Advising Note Encoder ($Z_{note}$)}
The input $N \in \{ N_1, N_2 .. N_t \}$ here $t$ represents time, which is a document sequence. At first we perform BERT pre-trained embedding fine-tuning as proposed in \cite{DevlinCLT19}. First, we consider, each of the document consists of a sequence of sentences and each sentence has been considered as a sequence of words. We represent each of the sequence of word separated by token [CLS] while each sequence of sentence has been separated by [SEP] token as described in \cite{DevlinCLT19} proposed method. Then we map the final tokenized document into a sequence of input embedding vectors, one for each token, constructed by summing the corresponding word, segment, and positional embeddings, thus it is called input representation vector. Now, we use multi-layered bidirectional Transformer encoder (BERT) \cite{DevlinCLT19} pre-trained embedding to map input representation vectors into a sequence of contextual embedding vectors. Then, the sequence of contextual embedding vectors are passed through a Bidirectional LSTM (BiLSTM) \cite{ZhangYZYZ20}. The BiLSTM layer concatenates the outputs from 2 hidden layers of opposite direction to the same output and can capture long term dependencies in sequential text data. The maxpooling layer takes the hidden states of the BiLSTM layer as input and outputs the final text representation $Z_{note}$ \cite{ZhangYZYZ20}.

\subsubsection{Student Spatiotemporal Information Representation ($Z$)}
The final students' spatiotemporal information representation $Z$ is obtained by concatenating the representations of sequential advising note, temporal student performance, along with static student demographic information. The representation of each student is $z_p \in Z = [ Z_{temporal}, Z_{static}, Z_{note} ]$ the size of this vector is $d_{temporal}, d_{static}, d_{note}$.

\subsection{Multi-Task Neural Cascade Networks}
We leverage the final task $L$ as hierarchical composition of five tasks ($L \in \{ L_1, L_2, L_3, L_4, L_5 \}$) for future dropout, type of dropout, next semester dropout, duration of dropout and cause of dropout tasks respectively, to train our student retention risk predictor by developing a Multimodal Spatiotemporal Neural Fusion network for MTL (\emph{MSNF-MTCL}). We formulate two types of losses:

\begin{itemize}
  \item Categorical cross-entropy loss for classification task
  \begin{equation}
    L_i^{cat} = -( y_i^{cat} log(p_i) + (1- y_i^{cat}) (1-log(p_i) )
    \label{eqn:cat}
  \end{equation}
  where $p_i$ denotes probability of the classification task and $y_i^{fd}\in\{y_1,..,y_n\}$ denotes the ground-truth labels.
  
  \item Euclidean loss for regression task
  \begin{equation}
    L_i^{reg} = {|| \hat{y}_i^{reg} - y_i^{reg} ||}_2^2
    \label{eqn:reg}
  \end{equation}
  where $\hat{y}_i^{reg}$ is the continuous estimated regression task values and $y_i^{reg}$ is the ground truth.
\end{itemize}

We define each of task as of our multi-task model along with the final multi-source learning scheme as follows:

\subsubsection{Future Dropout (FD)}
This is a binary task involves predicting students' dropout in future (true/false) which is irrespective of the semester or duration. The learning objective is formulated as a two-class classification problem. For each sample, we use the cross-entropy loss $L_i^{1}$ similar to Eqn. \ref{eqn:cat} where where $p_i$ is probability of dropout in future and $y_i^{1}\in\{0,1\}$ denotes ground truth label.
\subsubsection{Type of Dropout (TD)}
This binary task aiming to further categorize dropout into temporary or permanent. Similar to Eqn \ref{eqn:cat}, we can formulate $L_i^{2}$ where where $p_i$ is probability of type of dropout (temporary dropout, permanent dropout and $y_i^{2}\in\{0,1\}$ denotes ground truth label.

\subsubsection{Next Semester Dropout (ND)}
This binary task aims to predict whether predicted dropped out student will be dropped out in next semester or not. We use the cross-entropy loss $L_i^{3}$ similar to Eqn. \ref{eqn:cat} where where $p_i$ is probability of next semester dropout and $y_i^{3}\in\{0,1\}$ denotes ground truth label.

\subsubsection{Duration of Dropout (DD)}
This regression task aims to predict how many semesters students survive if the dropout has been predicted. We use the Euclidean loss $L_i^{4}$ similar to Eqn. \ref{eqn:reg} where where $\hat{y}_i^{5}$ is the continuous estimated duration of dropout in terms of semester and $y_i^{5}$ is the ground truth.

\subsubsection{Cause of Dropout (CD)} This task aims to predict the causes of dropout, i.e. one of the 15 causes as stated in Table~\ref{tab:features}. We use the cross-entropy loss $L_i^{5}$ similar to Eqn. \ref{eqn:cat} where where $p_i$ is probability of each cause of dropout and $y_i^{5}\in\{0,1,..,14\}$ denotes ground truth label.

\subsubsection{Multi-Conditional Training}
We employ five different tasks on our encoded students' information space $Z$, there are different types of labels in each training sample. While training on the samples, we follow the hierarchy of $L_1(FD)\rightarrow L_2(TD) \rightarrow L_3(ND) \rightarrow L_4(DD) \rightarrow L_5(CD)$ and develop an overall learning target as follows
\begin{equation}
  L(\Theta) = L_1 + L_2 + L_3 + L_4 + L_5
  \label{eqn:all}
\end{equation}
While computing $L$, we abide the following strategies: if $y_i^{1}=0$ (no dropout), then we set, $L_2 = L_3 = L_4 = L_5 = 0$, if $y_i^{1}=1$ (no dropout) and $y_i^{2}=0$ (permanent dropout), then we set, $L_3 = L_4 = L_5 = 0$, , if $y_i^{1}=1$ (no dropout), $y_i^{2}=0$ (permanent dropout), and $y_i^{3}=0$ (next semester dropout = true), then we set, $ L_4 = L_5 = 0$. We compute $L$ considering altogether as per Eqn. \ref{eqn:all} for all other cases.
\section{Experiments}

\begin{table*}[!h]
 \begin{center}
\begin{small}
 \caption{Comparison of \emph{MSNF-MTCL} performance on our dataset with different baseline models}
 \label{tab:result_table_1}

 \begin{tabular}{|p{1.8cm}|p{1.6cm}|p{1.45cm}|p{1.45cm}|p{1.45cm}|p{1.45cm}|p{1.45cm}|}

  \\ \hline
 {\bf Data \#Classes}   & {\bf B1} & {\bf B2} & {\bf B3} & {\bf V1}& {\bf V2} & {\bf V3 (Ours)} \\
 \hline

  FD (2) & $72.45\pm9.3$ & $73.51\pm8.6$ & 75.47$\pm$6.8 & $80.76\pm3.8$ & $82.84\pm4.2$ & {\bf 98.78$\pm$0.01} \\ \hline
  TD (2) & $61.65\pm8.8$ & $66.56\pm9.1$ & 70.77$\pm$8.4 & $76.42\pm4.3$ & $79.54\pm5.3$ & {\bf 89.73$\pm$0.01} \\ \hline
  ND (2) & $60.65\pm10.4$ & $65.63\pm9.2$ & 68.84$\pm$8.3 & $78.73\pm4.2$ & $80.25\pm4.1$ & {\bf 93.25$\pm$0.01} \\ \hline
  DD & $5.35\pm0.87$ & $3.65\pm0.66$ & 2.3$\pm$0.56 & $1.1\pm0.18$ & $0.85\pm0.05$ & {\bf 0.045$\pm$0.002} \\ \hline
  CD (15) & $59.83\pm8.4$ & $59.42\pm9.3$ & 60.27$\pm$11.53 & $68.54\pm5.3$ & $70.54\pm3.5$ & {\bf 85.53$\pm$0.02} \\ \hline

 \end{tabular}
\end{small}
 \end{center}
\end{table*}
\subsection{Baseline Models}
Since, multi-task multi-modal neural fusion on educational dataset is a novel problem for student retention risks estimation, we could not find state-of-art solutions that match with our problem as a baseline. In this regard, we implement few nearest problems along with their solutions and formulate similar problem using our proposed \emph{MSNF-MTCL} framework. Apart from that, to establish the importance of different modules of our framework, we develop different versions of \emph{MSNF-MTCL} consist of different combinations of proposed modules. The baselines and different versions of \emph{MSNF-MTCL} framework have been described below:
\begin{itemize}
  \item {\bf B1 (Jayaraman Model) \cite{drop6}:} This framework utilized only advising note and proposed a lexicon-based sentiment analysis technique to extract features and applied SVM machine learning techniques on the features to predict student dropout. The framework utilized Bing Lexicon \cite{Liu10} model for feature extraction that consists of 6,800 words, 2,000 positive and 4,800 negative sentiments.
  \item {\bf B2 (Pellagatti Model) \cite{drop5}:} This framework considered students' static and students' temporal structured data towards building a generalized mixed-effects random forest (GMERF).
  
  \item {\bf B3 (Single Task Fusion and Replacing BERT with Doc2Vec) \cite{ZhangYZYZ20}:} This framework is the closest one to our solution that has been developed to predict mortality of patients from electronic health records (EHR). It followed a spatiotemporal neural fusion of patient notes, patients' static demographic data and patient's temporal hospital information altogether into a fused layer that has been utilized to solve single task, predicting patients' mortality. Instead of using lexicon tokenization and BERT model for encoding patient notes, this framework utilized Doc2Vec embedding \cite{LeM14}.
  
  \item {\bf V1 (MSNF-MTCL with Structured Data Only):} This is a version of our proposed core \emph{MSNF-MTCL} model where we completely removed Temporal Advising Notes input and considered only Structured data i.e. Temporal Student Performance and Static Student Description inputs along with their encoders.
  
  \item {\bf V2 (MSNF-MTCL with Unstructured Advising Notes Only):} This is a version of our proposed core \emph{MSNF-MTCL} model where we included only Temporal Advising Notes input and its corresponding encoder.
    
  \item {\bf V3 (MSNF-MTCL):} This is a complete \emph{MSNF-MTCL} model including all modules and inputs.
\end{itemize}

\begin{table*}[!htbp]
 \begin{center}
\begin{small}
 \caption{Bias detection and mitigation experiment results. Here, column represents bias detection metrics: Statistical Parity Difference (SPD), Equal Opportunity Difference (EOD), Average Odds Difference (AOD) and Disparate Impact (DI); while rows represent bias mitigation techniques: Reweighing (RW), Adversarial Debiasing (AB), Reject Option Based Classification (ROBC), Equalized odds post processing (EOPP), Disparate impact remover (DIR), Learning fair representation (LFR), Calibrated equalized odds postprocessing (CEOP) and Prejudice remover (PR)} for each of the task: future dropout (FD), next semester dropout (ND), type of dropout (TD), duration of dropout (DD) and cause of dropout (CD)
 \label{tab:bias_mitigation}

 \begin{tabular}{|p{2cm}|p{1.5cm}|p{1.5cm}|p{1.5cm}|p{1.5cm}|p{0.7cm}|p{0.7cm}|p{0.7cm}|p{0.7cm}|p{0.7cm}|}
 
  \hline
  & {\bf SPD}  & {\bf EOD} & {\bf AOD} & {\bf DI} & {\bf FD}& {\bf ND} & {\bf TD}& {\bf DD}& {\bf CD}\\
 \hline
   
  Fairness target & -0.1 to 0.1  & -0.1 to 0.1 & -0.1 to 0.1 & 0.8 to 1.2 & 98.78& 89.73& 93.25& 0.045& 85.53\\ \hline
  Initial& 0.25 & -0.18 & -0.19 & 0.53 & &&&&\\ \hline
  RW & {\bf 0.05} & {\bf -0.03} & -0.15 & {\bf 0.95} & 91.85& 86.73& 90.55& 0.223& 81.47\\ \hline
  AB & 0.09 & {\bf -0.07} & -0.11 & {\bf 1.0} & 90.34& 87.45& 89.65& 0.23& 81.34\\ \hline
  ROBC & {\bf 0.06} & -0.11 & {\bf 0.08} & {\bf 0.91} & 91.24& 86.43& 89.38& 0.09& 80.44\\ \hline
  EOPP & 0.18 & -0.15 & {\bf -0.07} & {\bf 0.88} & 90.75& 87.47& 85.76& 0.23& 80.43\\ \hline
  DIR & {\bf 0.06} & {\bf -0.09} & -0.11 & {\bf 0.11} & 88.36& 85.83& 86.99& 0.24& 83.05\\ \hline
  LFR & 0.20 & -0.10 & {\bf 0.01} & {\bf 1.0} & 90.77& 83.84& 88.87& 0.145& 82.75\\ \hline
  CEOP & {\bf 0.05} & {\bf -0.05} & -0.11 & {\bf 0.89} & 89.76& 85.4& 90.93& {\bf 0.049}& 80.34\\ \hline
  PR & {\bf 0.06} & {\bf -0.09} & {\bf -.04} & {\bf 0.91} & {\bf 93.53}& {\bf 88.83}& {\bf 92.54}& 0.055& {\bf 83.46}\\ \hline
 \end{tabular}
\end{small}
 \end{center}
\end{table*}

\subsection{Results}
We considered accuracy $accuracy = \frac{TP+TN}{TP+TN+FP+FN}\;where\; TP, TN, FP\;FN\;denote\;true-positive,true-negative,false-positive,false-negative$ and Standard Deviation $\pm\%$ as evaluation metric for classification tasks. We considered root mean squared deviation (RMSD) as evaluation metric for regression tasks. We implemented baseline algorithms and our framework using python-based Keras library. We train the model using a learning rate of 0.001 for 16k iterations, and 0.0001 for the next 5k until the training converges. We train the model in 4 GPUs, each GPU holding 1 mini-batch (so the effective mini-batch size is x4).

While developing baseline algorithms, we designed 5 single task models for 5 retention risks. We considered 75\% of students' data as training and rest of 25\% of students' data as testing data during training and similar experiment has been conducted 10 times on 10-fold cross experiment to generate the results. We also utilized Synthetic Minority Oversampling Technique (SMOTE) to correct the imbalance \cite{FinlayPC14}. SMOTE is a popular and robust technique that uses a combination of oversampling the minority class and undersampling the majority class which results in better classifier performance than just oversampling or undersampling. Table \ref{tab:result_table_1} shows detail results of our experiment and comparisons.

\begin{figure}
 \centering
 \includegraphics[width=\linewidth]{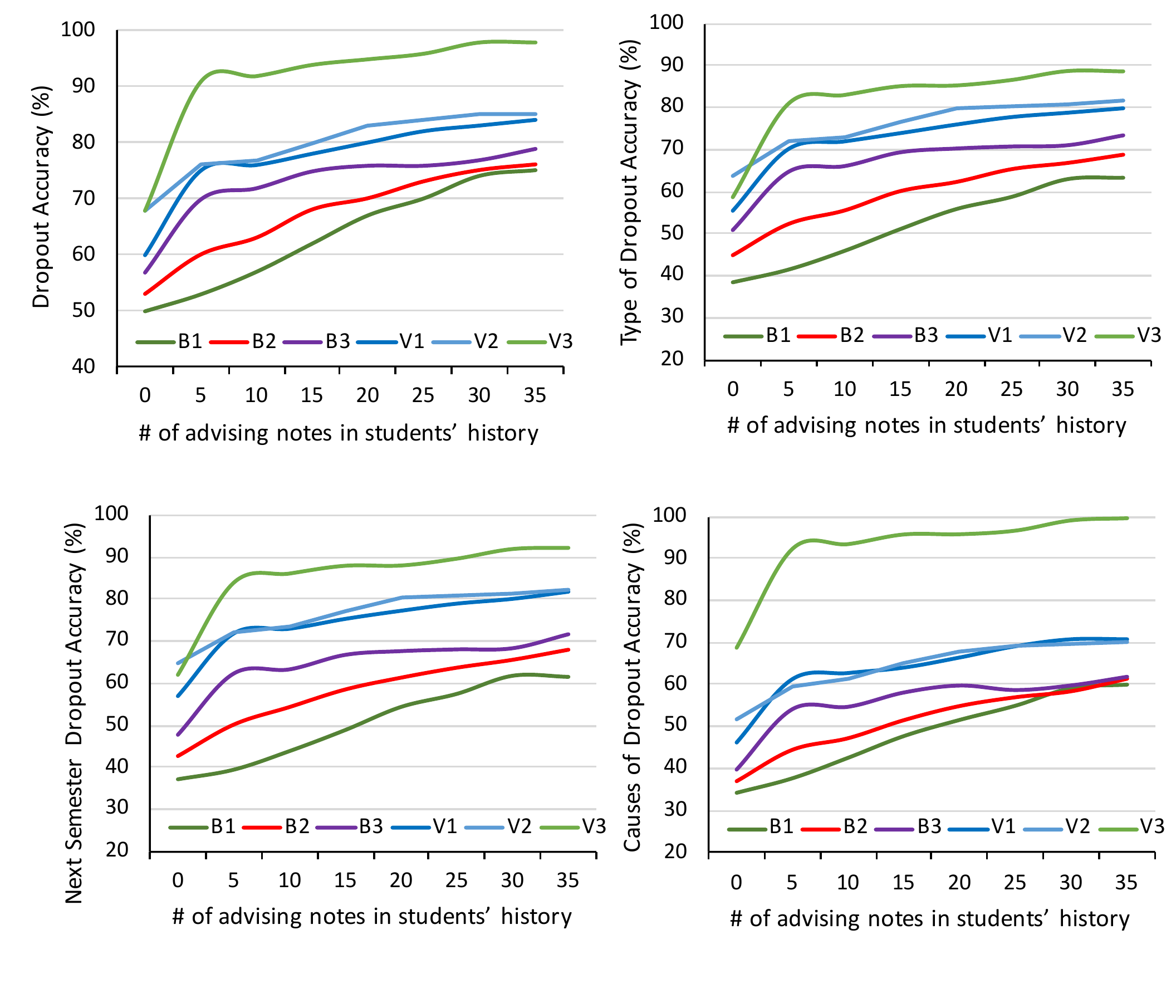}
 \caption{Accuracy changes of five different retention risk prediction tasks using our framework over number of available advising notes}
 \label{fig:advising_notes_results}
\end{figure}

\begin{figure}
 \centering
 \includegraphics[width=0.8\linewidth]{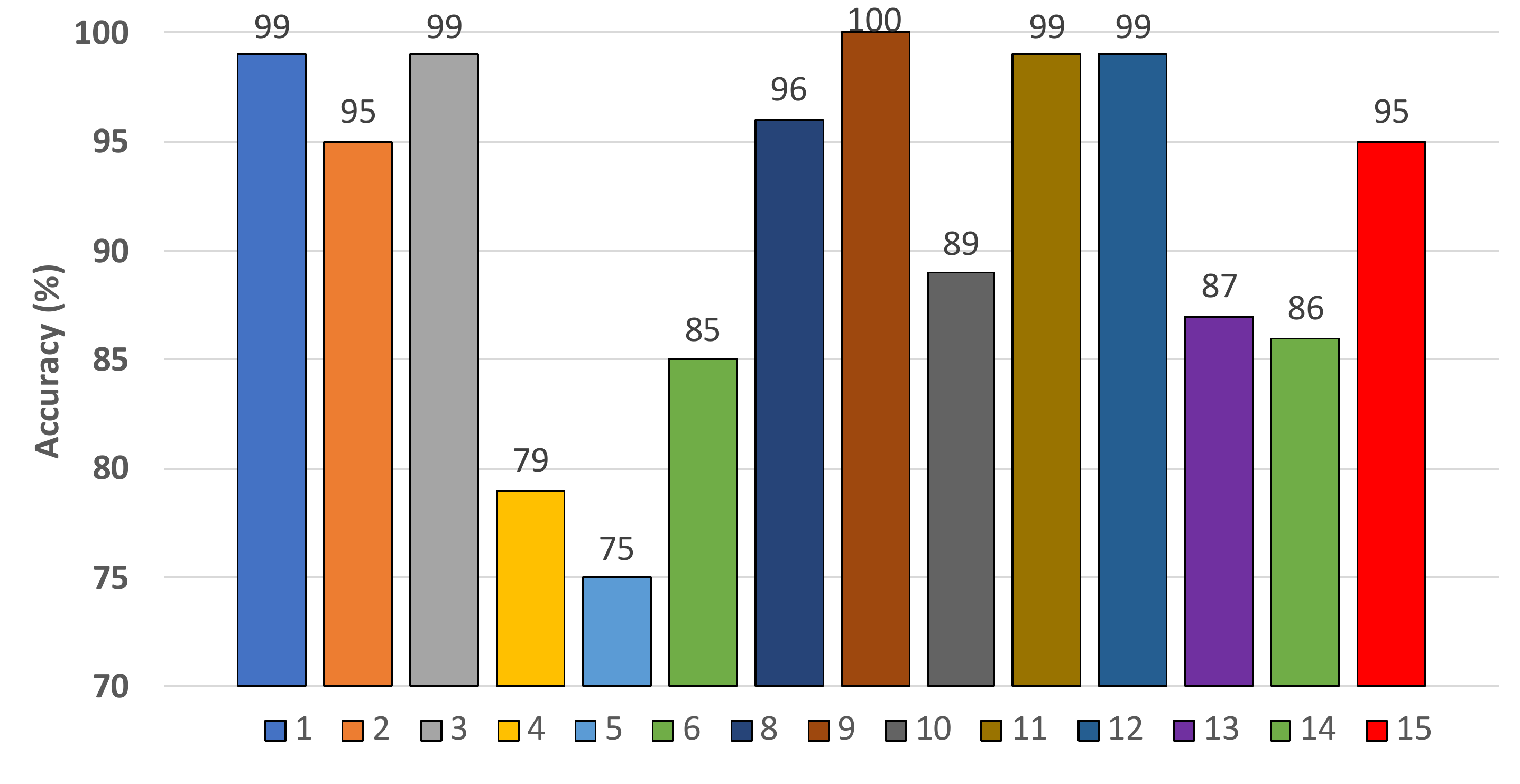}
 \caption{Causes of Dropout prediction results using our overall framework. The causes of dropout have been indexed with: 1. financial, 2. family, 3. marriage, 4. physically ill, 5. death of family member, 6. personal, 7. own death, 8. accident, 9. struggling with grades, 10. COVID-19 family death, 11. COVID-19 financial, 12. COVID-19 online class attending hardship, 13. internship, 14. traveling, 15. mentally ill. We removed 7.own death due to ethical reason.}
 \label{fig:causes_dropout_result}
\end{figure}

In Table \ref{tab:result_table_1}, we clearly can see that our proposed method (V3-Ours) perform better than any other baseline frameworks (B1, B2 or B3) for all student retention risk classification/estimation. If we take closer look, we can see that, utilizing only advising notes (V1) and only structured data (V2) versions of our framework not only outperform their related baselines (only advising note B1 and only structured data B2), the outperform state-of-art single task spatiotemporal fusion model using Doc2Vec embedding framework which has been successfully applied on EHR data before.

\subsection{Bias Detection and Mitigation}
Table \ref{tab:description} shows that the data is biased in terms of gender (female-male ratio is 28\% by 72\%) which has potential threat to AI fairness in our model. We utilize IBM AI Fairness 360 (AIF360) tool to detect and mitigate biases for dropout prediction in terms of gender considering "Male" as privileged group \cite{aif360}. Table \ref{tab:bias_mitigation} shows AIF360 implemented 4 bias detection metrics, their corresponding fairness target metric ranges and 8 bias mitigation techniques generated bias detection metrics. The central notions in this method: (1) all bias mitigation techniques are not appropriate for every dataset; (2) to select right mitigation technique, the bias detection metrics should be fair under maximum metrics; (3) accuracy drop due to bias mitigation should be minimum. Table \ref{tab:bias_mitigation} shows the final result of our bias detection and mitigation test for student dropout (only the first task of our multi-task model) where we can see that "Prejudice remover" technique provides maximum fairness (fair in 4 bias detection metrics) and least accuracy drop (accuracy drop of 3.33\%). Similarly, we can show that

\subsection{Discussion}
Fig. \ref{fig:advising_notes_results} illustrates the changes of accuracies over the number of availability of each student's advising note while predicting their retention risks (five different tasks) which clearly shows that different versions of our method (V1, V2 and V3) outperform baseline methods significantly in any number of advising notes' availability. Also, it can be clearly stated that, the prediction accuracy of each task increases as the number of available advising notes increases for each student in the testing data. Fig \ref{fig:causes_dropout_result} illustrates the prediction accuracies of individual dropout cause (15 dropout causes) using our proposed model, where we can see that (we removed cause "Own Death with index 7" due to ethical reason), predicting dropout due to financial condition, family reason, marriage related, struggling with grades, COVID-19 related financial, COVID-19 related struggling in attending online classes and mentally ill, are extremely accurate (95\%+). However, it has been extremely difficult to predict physical illness, death of family member and personal problem related college dropout from the educational data.

\subsection{Limitations and Future Work}
We utilized a large scale educational data of 18 years from only one university which may create distribution biases. To address biases, we additionally tested our framework for bias mitigation. Moreover, our reproduction of baseline models and evaluation on our dataset provide ample proof that our model outperforms baseline frameworks. In our framework, lower level cascaded task depends on the performance of upper level tasks' classification performances that we did not align with state-of-art models' implementations. The causes of dropout have been labeled in rolling basis, i.e., when a faculty advisor thought that current advisee needs to be assigned to a new cause, he reports to the system for an additional cause insertion. The administration officer review that cause and accept the inclusion request if that is absolutely valid. Our dataset consists of pre- and post- COVID-19 pandemic data. However, due to extremely poor number of data during post-COVID-19 era, we could not develop a new model to identify COVID-19 impacts on student dropout. In the current system, a faculty advisor can only assign a single cause for a single advising note, that made us difficult to predict multiple causes of a dropout incident which is common in real life case. In future, we aim to apply causal inference and information retrieval technique for facts finding to describe COVID-19 impacts and multiple causes extraction on student dropout more evidently. We also utilized pre-trained BERT embedding model that has been trained on wikipedia data. In future, we plan to develop a new embedding, "Educational BERT (EBERT)" trained on only educational advising notes to enhance efficiency of any student retention risk prediction.

\section{Conclusion}
Structured-unstructured data fusion in spatiotemporal domain across the educational institute has not been properly exploited by researchers due to the unavailability of such data and challenges of combining multi-modal educational signals. Our breakthrough approach that provides highest ever student dropout accuracy potentially can be adopted by educational policy makers and university management stakeholders in many other domains. Our novel problem formulation, a multi-task student retention risks estimation on 5 different student retention risk tasks, and solution, an efficient multi-task multi-modal spatiotemporal neural network model will open the door to many unsolved problems in educational data mining research. Moreover, the framework can be adapted in any databases in the world including employee, email, electronic health record or google search databases, and, can be utilized to solve extremely complex problems.

\end{document}